\documentclass[journal]{IEEEtran}

\usepackage[T1]{fontenc}
\usepackage{multirow}
\usepackage[table,xcdraw]{xcolor}
\usepackage{graphicx}
\usepackage{amsmath}
\usepackage{amsfonts}
\usepackage{algorithm}
\usepackage{amsthm,amssymb}
\usepackage{mathrsfs}
\usepackage{algorithmic}
\usepackage{makecell}
\usepackage{verbatim}
\usepackage{booktabs}
\setcellgapes{3pt}
\usepackage{amsthm,amsmath,amssymb,lipsum}
\usepackage{mathrsfs}
\setcellgapes{3pt}
\usepackage{multirow}

\usepackage[justification=centering]{caption}
\usepackage{color}
\usepackage[subfigure]{tocloft}
\usepackage{subfigure}
\usepackage{tabularx}
\usepackage{amsmath, bm}
\usepackage{authblk}
\usepackage{amsmath}
\usepackage{bm}
\makeatletter

\newcommand{\Rmnum}[1]{\expandafter\@slowromancap\romannumeral #1@}
\makeatother

\usepackage{hyperref} 
\hypersetup{
	hidelinks
}
\usepackage[numbers,sort&compress]{natbib}
\ifCLASSINFOpdf
\else
\fi
\usepackage{hyperref}

\begin{document}

\title{Visual Language Model based Cross-modal Semantic Communication Systems}

\author{Feibo Jiang, \textit{ Member, IEEE}, Chuanguo Tang, Li Dong, Kezhi Wang, \textit{Senior Member, IEEE}, Kun Yang, \textit{Fellow, IEEE}, Cunhua Pan, \textit{Senior Member, IEEE}
	\thanks{Feibo Jiang (jiangfb@hunnu.edu.cn) is with Hunan Provincial Key Laboratory of Intelligent Computing and Language Information Processing, Hunan Normal University, Changsha, China.
	
	Chuanguo Tang (202220294014@hunnu.edu.cn) is with School of Information Science and Engineering, Hunan Normal University, Changsha, China.
	
	Li Dong (Dlj2017@hunnu.edu.cn) is with Changsha Social Laboratory of Artificial Intelligence, Hunan University of Technology and Business, Changsha, China.
	
	Kezhi Wang (Kezhi.Wang@brunel.ac.uk) is with the Department of Computer Science, Brunel University London, UK.
	
	Kun Yang (kunyang@essex.ac.uk) is with the School of Computer Science and Electronic Engineering, University of Essex, Colchester, CO4 3SQ, U.K., also with Changchun Institute of Technology.
	
	Cunhua Pan (cpan@seu.edu.cn) is with the National Mobile Communications Research Laboratory, Southeast University, Nanjing 210096, China.
	
}
}


%
%

\markboth{Submitted for Review}%
{Shell \MakeLowercase{\textit{et al.}}: Bare Demo of IEEEtran.cls for IEEE Journals}
%

\maketitle

\begin{abstract}
Semantic Communication (SC) has emerged as a novel communication paradigm in recent years, successfully transcending the Shannon physical capacity limits through innovative semantic transmission concepts.  Nevertheless, extant Image Semantic Communication (ISC) systems face several challenges in dynamic environments, including low semantic density, catastrophic forgetting, and uncertain Signal-to-Noise Ratio (SNR). To address these challenges, we propose a novel Vision-Language Model-based Cross-modal Semantic Communication (VLM-CSC) system. The VLM-CSC comprises three novel components: 
(1) Cross-modal Knowledge Base (CKB) is used to extract high-density textual semantics from the semantically sparse image at the transmitter and reconstruct the original image based on textual semantics at the receiver. The transmission of high-density semantics contributes to alleviating bandwidth pressure.
(2) Memory-assisted Encoder and Decoder (MED) employ a hybrid long/short-term memory mechanism, enabling the semantic encoder and decoder to overcome catastrophic forgetting in dynamic environments when there is a drift in the distribution of semantic features. 
(3) Noise Attention Module (NAM) employs attention mechanisms to adaptively adjust the semantic coding and the channel coding based on SNR, ensuring the robustness of the CSC system.
The experimental simulations validate the effectiveness, adaptability, and robustness of the CSC system.
\end{abstract}

\begin{IEEEkeywords}
Semantic communication, knowledge base, vision language model, large language model, continual learning.
\end{IEEEkeywords}

\IEEEpeerreviewmaketitle

\section{Introduction}

As mobile communication technology has evolved from the first generation to the fifth generation, there has been a significant increase in transmission rates, approaching system capacities close to their limits \cite{1li2017intelligent}. In recent years, various emerging applications, such as the metaverse and virtual reality, have introduced substantial data streams \cite{2xu2022full}. Furthermore, these applications necessitate extensive connectivity over limited spectrum resources while demanding lower latency, posing significant challenges to conventional source-channel coding. Semantic Communication (SC) operates in the semantic domain by extracting the inherent meaning of data, eliminating redundant information, and achieving data compression while preserving its essential semantic content \cite{3yang2022semantic}.

With the rapid development of deep learning, many researchers have begun to explore end-to-end Image Semantic Communication (ISC) systems based on deep neural networks. For instance, ISC systems constructed using deep learning approaches such as Convolutional Neural Networks (CNN), Vision Transformers (ViT), and others have surpassed traditional solutions. Despite the significant achievements in the research of ISC based on deep learning, there remain some challenges:

\subsubsection*{1) Low semantic density}  Images are natural signals with
heavy spatial redundancy \cite{6he2022masked}. Traditional ISC systems directly encode the entire image, focusing on extracting low-level semantic information at the pixel level. However, text is a human-invented signal that possesses high semantic and information density. Summarizing image information through text can surpass the low-level pixel-level semantics and achieve a more sophisticated high-level semantic understanding of objects and scenarios. Moreover, traditional ISC systems lack the ability to leverage the interpretability of knowledge bases (KBs), resulting in a black-box model based on deep learning for the semantic encoder and decoder with limited explainability of semantics. 

\subsubsection*{2) Catastrophic forgetting} 
ISC systems often operate in dynamic environments, leading to a drift in the feature distribution of transmitted image data and channel state over time. Consequently, the real data distribution becomes inconsistent with the distribution during training, resulting in a decline in the performance of the semantic encoder and decoder.
Continual learning of the semantic encoder and decoder is necessary to improve the performance of the ISC system. However, during continual learning, the existing knowledge of the encoder and decoder may be disrupted or overwritten by new knowledge, leading to catastrophic forgetting in the learning process \cite{7zhang2022deep}. As a result, it becomes unable to adapt to semantic transmission in dynamic environments.
	
\subsubsection*{3) Uncertain Signal-to-Noise Ratio (SNR)} 
In wireless communications, traditional deep learning-based ISC systems typically consider a few discrete SNR conditions during the training phase, which cannot cover all possible SNR scenarios. As a result, the performance may severely degrade when there is a mismatch between the channel conditions during training and inference phases \cite{8bourtsoulatze2019deep}. Training the semantic/channel encoder and decoder with consideration for multiple SNR conditions and performing switching based on specific SNR values during the inference phase can lead to substantial storage and computational overhead \cite{9xu2021wireless}.

Vision Language Models (VLMs) with billions of parameters represent the latest advancements in the field of large AI models. Through extensive pre-training on vast amounts of data, these VLMs acquire rich language and visual knowledge, leading to significant breakthroughs in areas such as natural language processing and computer vision \cite{10zhao2023survey}. In ISC systems, VLMs demonstrate immense potential. Leveraging their capabilities in understanding and generating textual and visual content, VLMs enable more accurate semantic comprehension and semantic feature extraction, thereby offering a more intelligent and efficient ISC experience.
Therefore, we propose a novel VLM-based Cross-modal Semantic Communication (VLM-CSC) system to address the aforementioned challenges in ISC systems. Our contributions can be summarized as follows:

\subsubsection*{1) Cross-modal Knowledge Base (CKB)} 

We introduce a CKB, which consists of a Bootstrapping Language-Image Pre-Training (BLIP)-based KB at the transmitter for generating high-quality text descriptions consistent with images, and a Stable Diffusion (SD)-based KB at the receiver for reconstructing images matching the text descriptions. The text descriptions can be regarded as the extraction of high-level semantics from the images with low-level pixels, thereby enhancing the semantic density of the transmitted information. Additionally, these descriptions enable users to understand the extracted semantic content, thereby enhancing the explainability of the CSC system.

\subsubsection*{2) Memory-assisted Encoder and Decoder (MED)} 

We employ a MED to track changes in dynamic environments while avoiding catastrophic forgetting during the learning process. Specifically, we design a storage pool consisting of two types of memory: Short-Term Memory (STM) and Long-Term Memory (LTM). The STM is used to store the new data from the current environment, while the LTM stores historically significant data from previously encountered distributions. When training the CSC system, we input data from both the STM and LTM. This enables the semantic encoder and decoder to review all the knowledge from previously trained data with different distributions while learning from the new data. As a result, the CSC system can acquire encoding and decoding capabilities for the new data distribution without significantly compromising its performance on the previously trained data distribution, thus avoiding catastrophic forgetting.

\subsubsection*{3) Noise Attention Module (NAM)} 
We present a NAM to dynamically adjust semantic coding and channel coding based on different SNR conditions. Specifically, after each encoder and decoder layer, we employ an attention module to adjust the weights for different encoders and decoders according to the SNR values provided by the channel feedback. When the SNR is high, the NAM evenly allocates higher weights to the semantic encoder and decoder to improve the encoding and decoding quality of the semantic features. Conversely, when the SNR is low, the NAM assigns higher weights to the channel encoder and decoder, improving the channel coding to combat the intense channel noise. This design ensures that the semantic features maintain high robustness under varying SNR conditions.

The rest of this paper is structured as follows. Section II presents the related work, Section III introduces the system model, Section IV provides a detailed description of the proposed VLM-CSC system, Section V outlines the experimental setup and results, and Section VI concludes the paper. 

\section{Related work}

\subsection{Deep learning enabled ISC systems}
Deep learning techniques are commonly employed in the construction of encoders and decoders for ISC systems. In \cite{46bourtsoulatze2019deep}, a comprehensive SC system based on CNNs was initially introduced, showcasing superior performance in Peak Signal-to-Noise Ratio (PSNR) when compared to traditional compression algorithms. In \cite{47dai2022nonlinear}, a novel Nonlinear Transform Source-Channel Coding (NTSCC) for SC systems was proposed, which leveraged a Variational AutoEncoder (VAE) to map the source signal to the latent space, and executed nonlinear transformation and channel coding in the space. Additionally, \cite{48dong2022semantic} presented an innovative SC system incorporating Semantic Slice-Models (SeSM) to facilitate adaptable model resemblance under diverse requirements. Furthermore, \cite{49huang2022toward} introduced a Reinforcement Learning-based Adaptive Semantic Coding (RL-ASC) for image data. RL-ASC utilized a combination of VAE, RL, and generative adversarial networks (GANs) to encode, allocate, and decode semantic concepts. 

Although convolutional and ViT-based autoencoders have shown promising results, their feature extraction capabilities are limited compared to state-of-the-art VLMs. This limitation arises from constraints posed by model parameters and the availability of training data. 

\subsection{Vision language models}

VLMs are a class of large AI models capable of simultaneously processing both image and text information \cite{15cao2303comprehensive}. They find extensive application across various visual language tasks, encompassing image description, visual question answering, text-to-image generation, and other multimodal tasks. In \cite{17furst2022cloob}, a contrastive loss function was utilized to train both image encoders and text encoders. This loss function aimed to minimize the feature space distance between matching image-text pairs, enabling the learning of semantically relevant visual language features while reducing the dependence on large amounts of annotated data.
In \cite{18wang2021simvlm}, images were treated as prefixes in language models. They were decomposed into multiple blocks, concatenated with text sequences as input, and used to predict the subsequent parts of the text sequences.  Furthermore, in \cite{20tan2019lxmert}, a cross-attention mechanism was employed to integrate visual and language features. This mechanism allowed the two modalities to reference and enhance each other, facilitating the learning of more comprehensive and refined visual language features. The approach demonstrated applicability to various downstream tasks.

VLMs aim to understand the correlation between images and text, enabling accurate visual description or image generation. Future research involves deep integration of self-supervised pre-training techniques and VLMs. This integration will help extract cross-modal relationships between visual and language features, providing a stronger foundation for downstream tasks.

\subsection{Continual learning}
Continual learning can effectively mitigate the problem of catastrophic forgetting in dynamic environments. In \cite{51xu2023age}, the authors discuss continual learning in Mobile Edge Computing (MEC) networks, focusing on age-aware optimization for data selection and aggregator placement. They also present a prototype implementation involving diverse user equipment and cloudlets. 
In \cite{52yu2022continual}, the authors propose a continual learning digital predistortion algorithm for linearizing radio frequency power amplifiers in 6G wireless communications. The algorithm demonstrates effectiveness in adapting to both new and known operating states with low long-term complexity. 
In \cite{53zhang2022cross}, the authors address the challenge of forgetting tasks in cross-edge federated learning by preserving past knowledge through continual learning. They achieve enhanced accuracy across various tasks with minimal storage cost.
Furthermore, in \cite{54zhou2022continual}, the authors employ continual learning to enable adaptive downlink beamforming optimization in dynamic environments. The proposed approach addresses task mismatch and exhibits good adaptability with low complexity.

Recent advancements in continual learning have been directed towards more challenging scenarios, specifically those where task boundaries are unknown. In these contexts, researchers have focused on developing sample selection strategies to identify which samples should be stored in the buffer for model training. This approach aims to improve the efficiency and effectiveness of continual learning in handling unknown task boundaries.
\begin{figure*}[htpb]
	\centering
	\includegraphics[width=17cm]{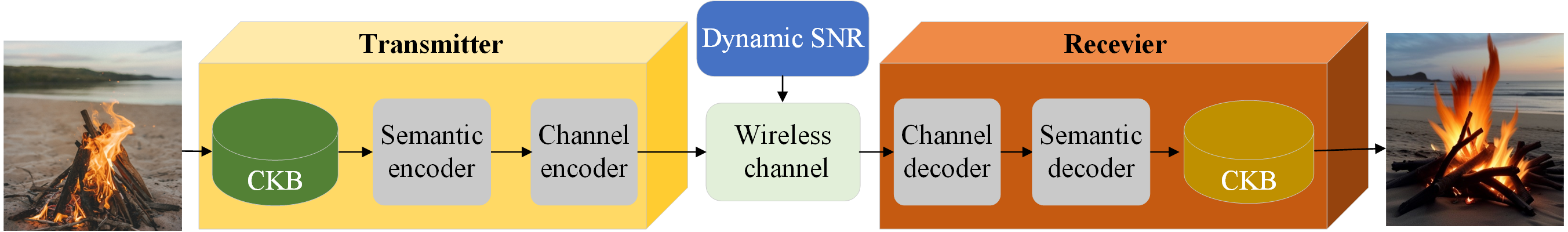}
	\caption{The system model of the CSC.}
	\label{fig:fig1}
\end{figure*}

\section{System model and problem formulation}
The considered CSC system consists of three components: a transmitter, a receiver, and a physical channel, as illustrated in \textbf{Fig. \ref{fig:fig1}}. The physical channel ensures the correct exchange of semantic information over the transmission medium with dynamic SNR.

\subsection{Transmitter}
The input to the transmitter is an image represented by the matrix $\bm{\mathrm{x}}\in \mathbb{R}^{H\times W \times C}$, whose size is $H(height)\times W(weight) \times C(channel) $. In the transmitter, the input image $\bm{\mathrm{x}}$ is mapped to symbols $\bm{\mathrm{y}}$ for transmission over the physical channel. The transmitter consists of three independent components: a CKB for cross-modal semantic extraction, a semantic encoder, and a channel encoder. The CKB is used to extract semantic information from the image and represent it as the corresponding textual information. The semantic encoder and channel encoder are responsible for semantic coding, and channel coding and modulation, ensuring that the encoded semantic information can be smoothly transmitted over the physical channel. The encoded symbol sequence $\bm{\mathrm{y}}$ can be represented as:
\begin{equation}\label{eq:shi1}
	 \bm{\mathrm{y}}=C_{\beta}(S_{\alpha}(K_{\theta}(\bm{\mathrm{x}}),\mu),\mu)
\end{equation}
where $K_{\theta}(\cdot)$ is the CKB with the parameter set $\theta$, $S_{\alpha} (\cdot)$ is the semantic encoder with the parameter set $\alpha$, and $C_{\beta}(\cdot)$ is the channel encoder with the parameter set $\beta$, $\mu$ is the channel SNR that can be estimated and fed back to the semantic encoder and channel encoder.

\subsection{Wireless channel}
The transmitter sends encoded symbols $\bm{\mathrm{y}}$, which is transmitted through the physical channel to the receiver. The channel output sequence $\bm{\mathrm{\hat{y}}}$ at the receiver can be expressed as: 
\begin{equation}\label{eq:shi2}
	\bm{\mathrm{\hat{y}}}=\bm{\mathrm{h}}\bm{\mathrm{y}}+\bm{\mathrm{n}}
\end{equation}
where $\bm{\mathrm{h}}$ represents the channel gain, and $\bm{\mathrm{n}}$ is Additive White Gaussian Noise (AWGN). 

\subsection{Recevier}
Similar to the transmitter, the receiver consists of three components: a channel decoder, a semantic decoder, and a cross-modal knowledge base for semantic reconstruction. The semantic decoder and channel decoder are used to decode textual information from received symbols, while the cross-modal knowledge base is employed for image reconstruction based on the corresponding textual information. The decoded image can be represented as:
\begin{equation}\label{eq:shi3}
	\bm{\mathrm{\hat{x}}}=K_{\theta'}^{-1}(S_{\delta}^{-1}(C_{\gamma}^{-1}(\bm{\mathrm{\hat{y}}},\mu),\mu))
\end{equation}
where $C_{\gamma}^{-1}(\cdot)$ is the channel decoder with the parameter set $\gamma$, $S_{\delta}^{-1}(\cdot)$ is the semantic decoder with the parameter set $\delta$ and $K_{\theta'}^{-1}(\cdot)$ is the cross-modal knowledge base with the parameter set $\theta'$. 

For the purpose of reconstructing image information from the semantic level, maintaining the consistency of textual semantics between $\bm{\mathrm{s}}$ and $\bm{\mathrm{\hat{s}}}$ is crucial. Here, $\bm{\mathrm{s}}=K_{\theta}(\bm{\mathrm{x}})$ represents the extracted textual semantic information from the image, and $\bm{\mathrm{\hat{s}}}=S_{\delta}^{-1}(C_{\gamma}^{-1}(\bm{\mathrm{\hat{y}}},\mu),\mu)$ represents the recovered textual semantic information after decoding. We utilize Cross-Entropy (CE) as the loss function:
\begin{equation}\label{eq:shi4}
	L_{CE}(\mathbf{s},\bm{\mathrm{\hat{s}}})=-\sum_{l=1}^{L}q(w_{l})\log(p(w_i))+(1-q(w_l))\log(1-p(w_i))
\end{equation}
where $q(w_l)$ denotes the real probability of the appearance of the $l$-th word $w_l$ in the sentence $\mathbf{s}$, and $p(w_l)$ represents the predicted probability of the appearance of the $l$-th word $w_i$ in the sentence $\bm{\mathrm{\hat{s}}}$. CE is employed to measure the difference between two probability distributions. By minimizing the CE loss, the semantic encoder and decoder can learn the word distribution $q(w_l)$ in the source sentence $\bm{\mathrm{s}}$, which represents the meaning of words in terms of grammar, phrases, and contextual information. Hence, the goal pf the CSC system is to determine the parameters of the semantic/channel encoder and decoder 
${\alpha}^{\ast}$, ${\beta}^{\ast}$, ${\delta}^{\ast}$ and ${\gamma}^{\ast}$ that minimize the expected distortion as follows:
\begin{equation}
({\alpha}^{\ast},{\beta}^{\ast},{\delta}^{\ast},{\gamma}^{\ast})=\mathop{\arg\min}\limits_{{\alpha},{\beta},{\delta},{\gamma}}\mathbb{E}_{p(\mu)}\mathbb{E}_{p(\mathbf{s},\bm{\mathrm{\hat{s}}})}[L_{CE}(\mathbf{s},\bm{\mathrm{\hat{s}}})]
\end{equation}
where ${\alpha}^{\ast}$ is the optimal semantic encoder parameters, ${\beta}^{\ast}$ is the optimal channel encoder parameters, ${\gamma}^{\ast}$ is the optimal channel decoder parameters, and ${\delta}^{\ast}$ is the optimal semantic decoder parameters. $p(\mathbf{s},\bm{\mathrm{\hat{s}}})$ represents the joint probability distribution of the $\bm{\mathrm{s}}$ and $\bm{\mathrm{\hat{s}}}$, and $p(\mu)$ represents the probability distribution of the SNR.

\section{The VLM-CSC system}

Compared to traditional KBs based on Deep Neural Networks (DNNs), Knowledge Graphs (KGs), and other approaches, utilizing VLMs to construct KBs has several advantages: (1) VLMs are large AI models with billions of parameters and powerful cognitive abilities concerning world knowledge. They excel in tasks related to understanding, expressing, and generating both visual and natural language data from the semantic level. (2) Unlike traditional methods that rely on manual rules or structure definitions to describe knowledge, VLMs have the ability to automatically learn and extract knowledge from data. This enables them to generate appropriate semantic information, reducing the risk of information loss or ambiguity. (3) In SC systems, the process of understanding and interpreting the generated results is crucial. VLMs have the ability to generate semantic information in a manner that is understandable to humans, enabling both parties in communication to have a more accurate understanding and interpretation of each other's intentions and expressions. 

In this section, we will provide the implementation details of the proposed VLM-CSC system, which is illustrated in \textbf{Fig. \ref{fig:fig2}} as follows:
\begin{figure*}[htpb]
	\centering
	\includegraphics[width=17cm]{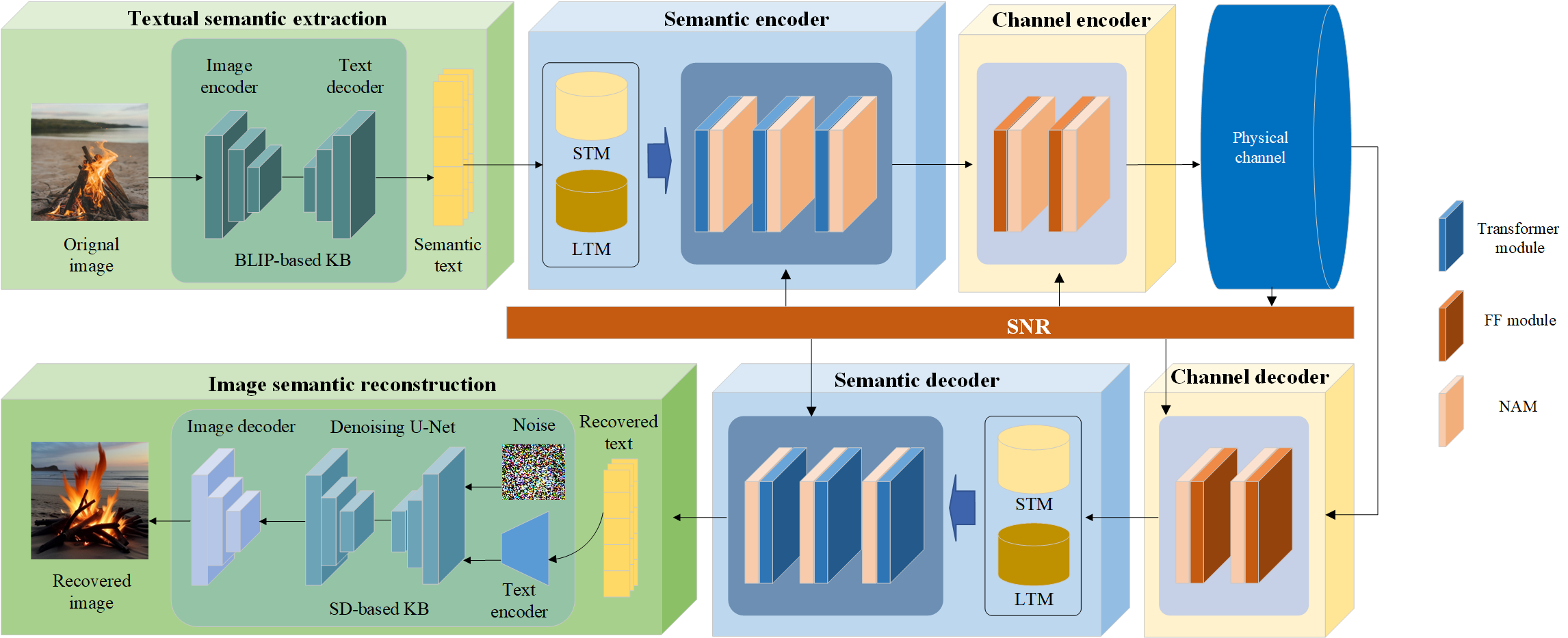}
	\caption{The proposed VLM-CSC system.}
	\label{fig:fig2}
\end{figure*}

\subsubsection{Textual semantic extraction}
To enhance the semantic density and interpretability of SC, a VLM called BLIP is employed at the transmitter to construct the CKB. 
The CKB encompasses a series of visual and language-related knowledge components. We employ the image encoder and text decoder from this CKB to perform cross-modal semantic extraction, thereby transforming the original image with low semantic density into a corresponding textual description with high semantic density.
For example, through cross-modal semantic extraction, the original image in \textbf{Fig. \ref{fig:fig2}} is transformed to the textual description "A fire is burning on a beach near the water".

\subsubsection{Semantic encoder and decoder}
The generated textual information from the CKB then proceeds to the semantic encoder. The semantic encoder consists of alternating transformer encoder layers and NAMs. The transformer encoder layers analyze and transform the textual information into a compact semantic representation. NAMs allow the semantic encoder to optimize the encoding process and maintain reliable semantic transmission, even in the presence of varying channel conditions.
At the receiver, the semantic decoder is composed of alternating transformer decoder layers and NAMs, with a structure opposite to that of the semantic encoder, aimed at reversing the semantic encoding process to recover the original textual information.

\subsubsection{Channel encoder and decoder}The encoded semantic features are passed through the channel encoder to undergo channel encoding and modulation, ensuring the effective transmission of semantic information over the physical channel. Similarly, the channel encoder also consists of alternating FeedForward (FF) layers and NAMs. At the receiver, the transmitted information through the physical channel is received and decoded using the channel decoder. To maintain information consistency, the channel decoder employs a structure opposite to that of the channel encoder.


\subsubsection{Image reconstruction} To facilitate a better understanding of the received textual information, we design a CKB for image reconstruction using a VLM called SD. 
The CKB encompasses a series of visual and language-related knowledge components. We employ the text encoder, the denoising U-Net and the image decoder from this CKB to perform image reconstruction.
Specifically, the textual information is first transformed into a conditional vector by the text encoder. Then, the denoising U-Net transforms the noisy image to a latent image feature vector aligning with the conditional vector. Finally, the latent image feature vector is processed by the image decoder to generate the final reconstructed image.

\subsubsection{Memory-assisted continual learning}
During the training phase of the VLM-CSC system, the latest samples are stored in an STM. When the STM becomes full, a kernel method is employed to select representative short-term samples to be transferred to an LTM. Then, the STM is emptied to buffer new samples in the next round.
The encoder and decoder sample from both STM and LTM during the training stage, thereby avoiding catastrophic forgetting. This approach ensures that the semantic encoder and decoder can access both recent and past information, allowing for continual learning and retention of previously learned knowledge.

\subsubsection{Training process of the VLM-CSC system}
Remarkably, BLIP and SD-based CKBs are pretrained VLMs that do not need to be trained specifically for the CSC system. The training process unfolds as follows: 
\begin{itemize} 
\item Joint training of channel encoder and decoder with NAMs: The channel encoder/decoder and NAMs are initially trained together by MED. This involves optimizing the parameters of these modules by minimizing the mutual information, which eliminates noise or fading effects during transmission and prevents signal distortion \cite{jiang2023large}. Then, the parameters of the channel encoder/decoder and NAMs are frozen. This ensures that their learned representations are preserved in subsequent training steps.

\item Joint training of semantic encoder and decoder with NAMs: The semantic encoder/decoder and NAMs are then trained by MED. The focus is on optimizing the parameters of these modules to minimize the loss between the original textual information and the reconstructed textual information. Eq. (\ref{eq:shi4}) can be applied as the loss function. Then, the parameters of the semantic encoder/decoder and NAMs are frozen to maintain the learned semantic representations.

\item Crossover-based iterative training: The training process iterates between the channel encoder/decoder and noise modules, and the semantic encoder/decoder and noise modules. This iteration continues until convergence of the entire VLM-CSC system is achieved.

\end{itemize} 

Next, we will provide a detailed explanation of each contribution in this paper.

\subsection{BLIP-based CKB for semantic extraction}
\begin{figure}[htpb]
	\centering
	\includegraphics[width=8cm]{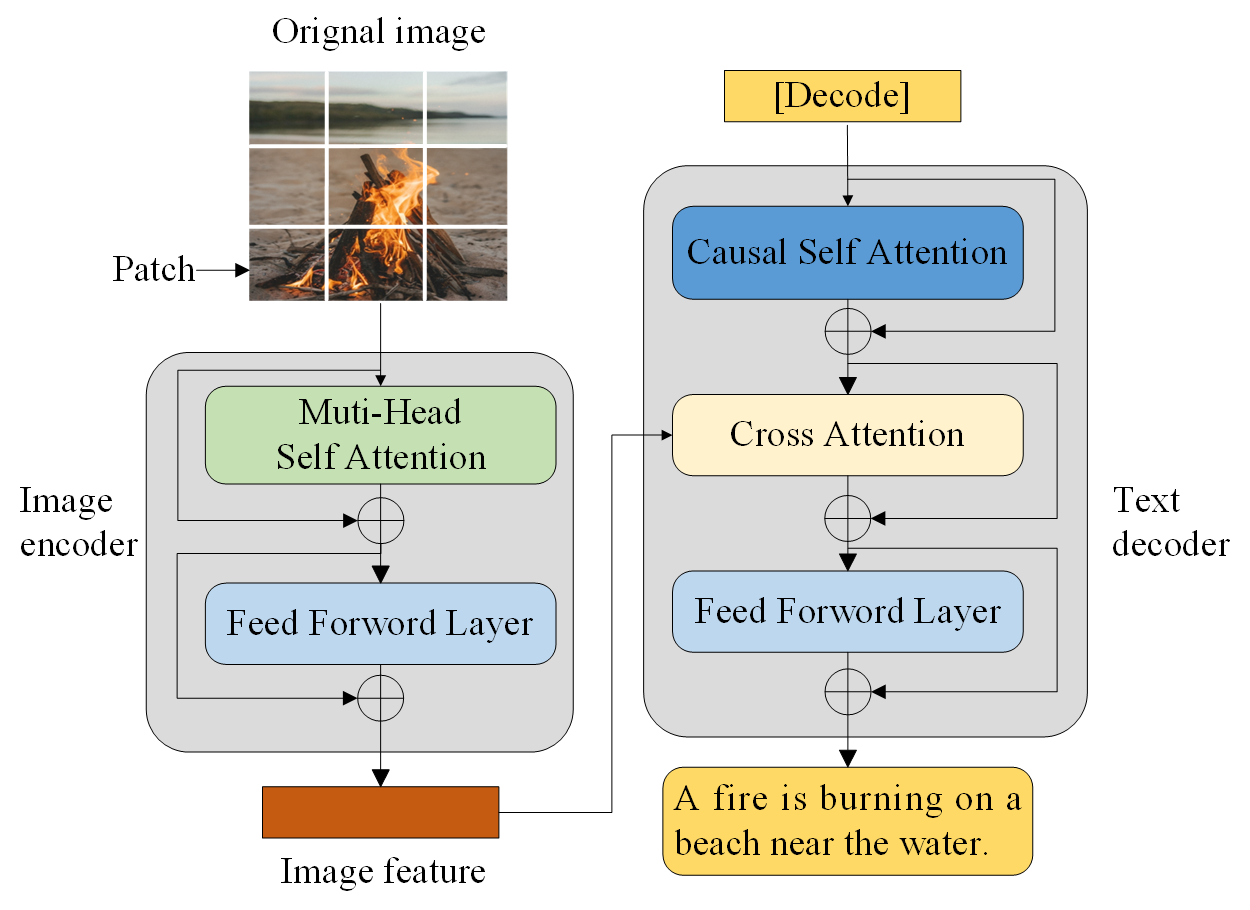}
	\caption{The architecture of BLIP-based CKB.} 
	\label{fig:fig3}
\end{figure}

The BLIP model, introduced by Salesforce AI Research, is a sophisticated VLM designed for understanding and generating content that involves both visual and textual elements \cite{30li2022blip}. The BLIP model possesses rich visual-linguistic knowledge and utilizes multiple knowledge components such as text encoders, image encoders, and image-grounded text decoders and decoders to perform various visual-linguistic tasks, such as image captioning, visual question answering, and multimodal classification. At the transmitter, we employ the BLIP model to construct the CKB and utilize the image encoder and image-grounded text decoder (abbreviated as text decoder) in the CKB to transform original image data into detailed textual descriptions containing image semantic information.
The workflow of the BLIP-based CKB is illustrated in \textbf{Fig. \ref{fig:fig3}}.

For a given image $\bm{\mathrm{x}}$, the process of extracting semantic information from image data and generating textual representation $\bm{\mathrm{s}}$ is as follows:

\subsubsection{Image encoder} The image encoder incorporates a feature extraction module based on the ViT. This module divides the input image into smaller patches and encodes each patch. Through multiple encoder layers with Multi-head Self-Attention (MSA) and FF sublayers \cite{42vaswani2017attention}, these patch vectors undergo processing to generate the textual representation of the image, which corresponds to the image features.


Initially, the image $\bm{\mathrm{x}}$ is segmented into a patch sequence $\bm{\mathrm{x}}_p$. Each patch represents a fixed-size image region in \textbf{Fig. \ref{fig:fig3}}. Subsequently, these patch sequences are fed into the image encoder to extract visual features from the image. The specific workflow of the image encoder is as follows:

 \begin{itemize} 
 	 \item MSA sublayer: the MSA layer allows the vector of each patch to interact with vectors of all other patches, capturing both global and local information in the image. The output of the MSA layer in the first image encoder layer can be calculated as follows:
\begin{equation}\label{eq:shi9}
	\bm{\mathrm{m}}_{msa,1}=\mathrm{MSA}(\mathrm{LN}(\bm{\mathrm{x}}_{p}))+\bm{\mathrm{x}}_{p}
\end{equation}
where $\bm{\mathrm{x}}_p$ is the $p$-th patch, $\mathrm{MSA}$ is the multi-head self-attention operator \cite{42vaswani2017attention} and $\mathrm{LN}$ is the layer normalization operator in ViT \cite{42vaswani2017attention}.
\item FF sublayer: The FF layer comprises linear layers and activation functions, facilitating non-linear transformations of vectors for each patch to enhance the model's adaptability. The output of the FF layer in the first image encoder layer is 
\begin{equation}\label{eq:shi10}
	\bm{\mathrm{m}}_{ff,1}=\mathrm{GeLU}(\mathbf{W}_{b,f} \cdot \mathrm{LN}(\bm{\mathrm{m}}_{msa,1})+\mathbf{b}_{b,f})+\bm{\mathrm{m}}_{msa,1}
\end{equation}
where $\mathbf{W}_{b,f}$ and $\mathbf{b}_{b,f}$ are the weights and biases of the FF layer in the image encoder of the BLIP model, and GeLU denotes the activation function.
 \end{itemize} 

Finally, the output of the image encoder with $L$ encoder layers is
 \begin{equation}\label{eq:shi10_1}
	\bm{\mathrm{m}}_{L}=\mathrm{LN}(\bm{\mathrm{m}}_{ff,L})
\end{equation}
where $\bm{\mathrm{m}}_{ff,L}$ means the output of the $L$-th encoder layer. 

\subsubsection{Text decoder} The text decoder of the BLIP model adopts a BERT structure, capable of generating image-related textual content, such as descriptions, titles, and dialogues, based on features extracted from images.
The text decoder is composed of multiple stacked decoder layers, each decoding layer comprising three sublayers: Causal Self-Attention (CSA), Cross Attention (CA), and FF sublayers. The specific workflow of the text decoder is as follows:
 \begin{itemize} 

\item CSA sublayer: 
CSA is a type of self-attention mechanism that only allows the attention model to access the current and previous inputs, but not the future inputs \cite{43yang2021causal}.
To ensure the causality of the textual generation process, the CSA sublayer utilizes a mask matrix to prevent the current token from accessing information from future tokens. Here, a token refers to the basic unit in the text, typically a word or a subword. The output of the CSA sublayer in the first text decoder layer is
\begin{equation}\label{eq:shi91}
	\bm{\mathrm{k}}_{csa,1}=\mathrm{CSA}(\mathrm{LN}(D_0))+D_0
\end{equation}
where $\mathrm{CSA}$ is the causal self-attention operator \cite{43yang2021causal}, $D_0$ is the initial token, which is typically set as "[Decoder]" by default.

\item CA sublayer: CA allows the vector of each token to interact with the feature vectors of visual information from the input image \cite{44chen2021crossvit}.  The output of the CA sublayer in the first text decoder layer can be calculated as follows:
\begin{equation}\label{eq:shi92}
	\bm{\mathrm{k}}_{ca,1}=\mathrm{CA}(\mathrm{LN}(\bm{\mathrm{k}}_{csa,1}),\bm{\mathrm{m}}_{L})+\bm{\mathrm{k}}_{csa,1}
\end{equation}
where  $\mathrm{CA}$ is the cross attention operator \cite{44chen2021crossvit}.
\item FF sublayer: The FF layer comprises linear layers and activation functions. The output of the FF layer in the first text decoder layer is 
\begin{equation}\label{eq:shi10}
	\bm{\mathrm{k}}_{ff,1}=\mathrm{ReLU}(\mathbf{W}'_{b,f} \cdot \mathrm{LN}(\bm{\mathrm{k}}_{ca,1})+\mathbf{b}'_{b,f})+\bm{\mathrm{k}}_{ca,1}
\end{equation}
where $\mathbf{W}'_{b,f}$ and $\mathbf{b}'_{b,f}$ are the weights and biases of the FF layer in the text decoder of the BLIP model, and ReLU denotes the activation function.
 \end{itemize} 


The final layer of the decoder transforms the output (via a linear projection and a softmax function) to predict the next token in the sequence. This output text is then used as an input for the next time step during the generation process until the final textual description $\bm{\mathrm{s}}$ of the image is produced.

\subsection{SD-based CKB for image reconstruction}
The SD model is an elaborate VLM collaboratively developed by Stability AI, which possesses rich visual-linguistic knowledge and is applicable to diverse tasks such as text-to-image and image-to-image generation  \cite{31rombach2022high}. At the receiver, we use the SD to construct the CKB and utilize the text-to-image components in the CKB to reconstruct images. The semantic reconstructor is composed of a text encoder, a feature generator, and an image decoder. 

\begin{figure*}[htpb]
	\centering
	\includegraphics[width=17cm]{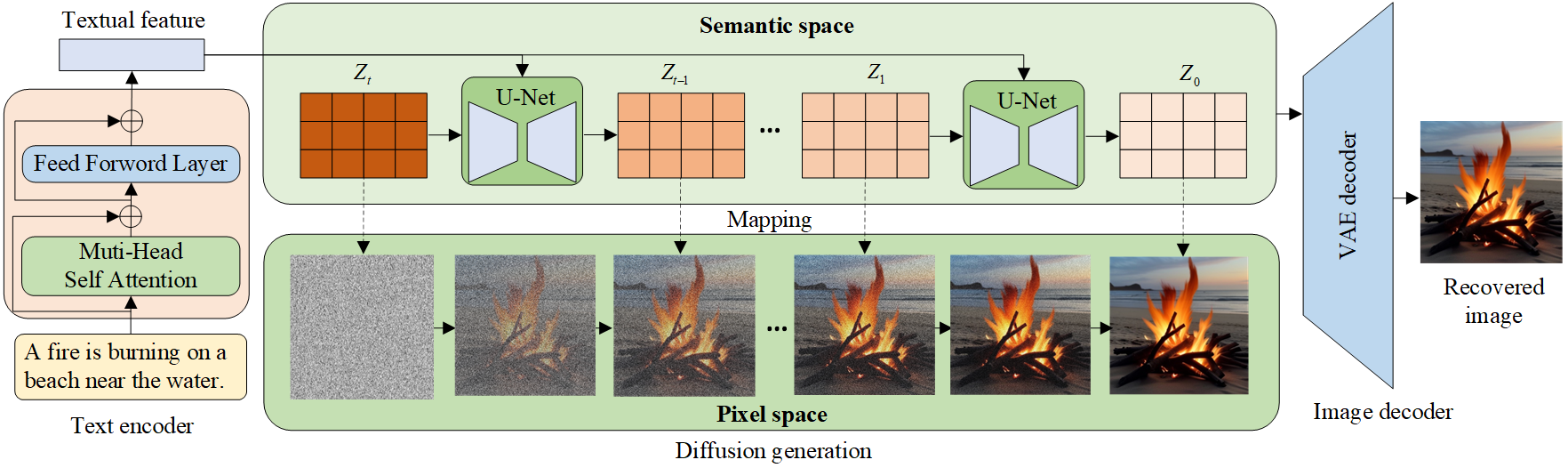}
	\caption{The architecture of SD-based CKB.} 
	\label{fig:fig4}
\end{figure*}

For a given semantic text $\bm{\mathrm{\hat{s}}}$, the image reconstruction process through the SD model is illustrated in \textbf{Fig. \ref{fig:fig4}} and is described as follows: 

\subsubsection{Text encoder} Text encoder is applied to transform the input text sequence into a semantic vector of fixed dimensions, serving as a control condition for the image feature generator. 
The text encoder is composed of multiple stacked encoding layers, each containing two sub-layers: MSA and FF. The residual connection and layer normalization are applied before each sublayer. This structure is similar to the image encoder in the BLIP model.

The input to the text encoder is the sequence $\bm{\mathrm{\hat{s}}}$ composed of words. Initially, each word is mapped to a fixed-length vector by word embeddings. These word embeddings, serve as the input to the text encoder. The encoder iteratively performs MSA and FF operations, ultimately producing a sequence composed of textual feature vectors. 

\subsubsection{Feature generator} 
An initial image feature vector composed of pure noise is input into the image feature generator. Textual feature vectors are injected into the noised feature vector to guide the noise removement. Through multiple iterations, noise is progressively removed, and an image feature vector containing textual information is obtained. The denoising step employs a U-Net structure, which adopts a CNN-based encoder-decoder structure to preserve spatial information while generating image semantic information. The iterative process of the image feature generator can be described by the following formula:
\begin{equation}\label{eq:shi101}
\mathbf{Z}_{t-1}=\frac{1}{\sqrt{{\alpha}}_t}(\mathbf{Z}_t-\frac{1-{\alpha}_t}{\sqrt{1-\overline{\alpha}_t}}f_{\theta}(\mathbf{Z}_t,t,\mathbf{d}))+{\sigma}_t\mathbf{Y}
\end{equation}
where $\mathbf{Z}_t$ represents the image feature vector at the time step $t$, ${\alpha}_t$ denotes the variance of the forward diffusion process, serving as a hyperparameter. $\overline{\alpha}_t=\prod_{i=1}^t{\alpha}_i$, $f_{\theta}$ represents the pre-trained noise prediction U-Net, $\mathbf{d}$ is the textual semantic vector, ${\sigma}_t\mathbf{Y}$ denotes the mean of the reverse diffusion process, where ${\sigma}_t=\sqrt{{1-\alpha}_t}$, and $\mathbf{Y}\sim \mathcal{N}(0,\mathbf{I})$ with $\mathbf{I}$ being the identity matrix.


\subsubsection{Image decoder} 
Due to the computational inefficiency of the diffusion operation, the denoising process of the image is performed in the compressed semantic space. Multiple iterations of denoising are conducted in the reduced semantic (feature) space, significantly improving the efficiency of image processing. Finally, we utilize the decoder of a Variational Autoencoder (VAE) to map the feature data in the semantic space back to the pixel space, reconstructing images that adhere to semantic consistency. As VAE learns the latent structure of a large amount of image data distribution, the decoder can provide more detailed information consistent with key semantics in the image by employing upsampling and interpolation during the decoding process, thereby enhancing the image quality in the pixel space.


\subsection{Memory-assisted encoder and decoder}
In dynamic environments, both the distribution of the transmitted contents and channel states will change over time. This necessitates that the CSC system continuously adjusts based on new input data and channel states to adapt to the evolving data distribution. However, such adjustments may lead to parameter updates in the encoder and decoder of the CSC system, potentially causing the catastrophic forgetting issue where old parameter values are overwritten or ignored \cite{7zhang2022deep}. Hence, continual learning diminishes the robustness of the encoder and decoder in the CSC system.

The memory-based learning strategy addresses the catastrophic forgetting problem in continual learning by diversifying the memorized content \cite{32ye2022continual}. We design a MED method with STM and LTM for both semantic encoder and decoder.  Below, we present the workflow of the MED as follows:

\begin{figure}[htpb]
	\centering
	\includegraphics[width=9cm]{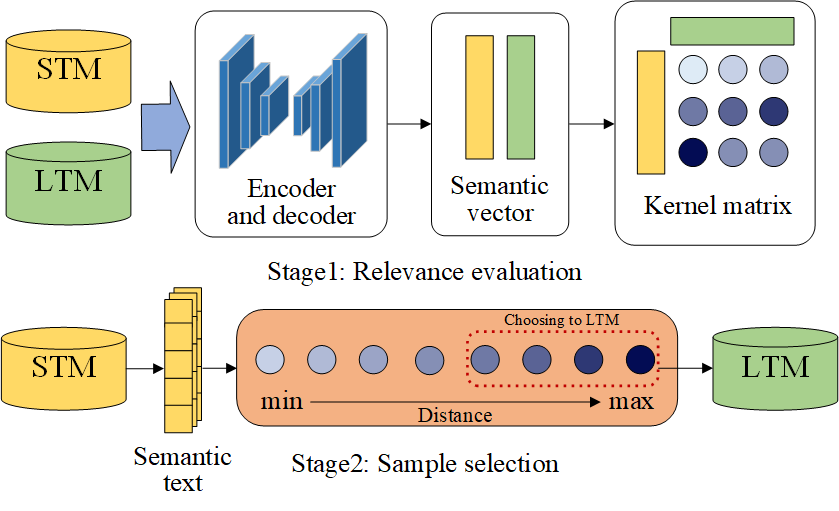}
	\caption{Memory-assisted encoder and decoder.}
	\label{fig:fig5}
\end{figure}

We denote $\mathcal{M}_{stm}=\{\bm{\mathrm{s}}_{i}^{stm}\}_{i=1}^{n_{stm}}$ and $\mathcal{M}_{ltm}=\{\bm{\mathrm{s}}_{j}^{ltm}\}_{j=1}^{n_{ltm}}$ as the sets representing dynamic samples stored in STM and LTM. $\bm{\mathrm{s}}_{i}^{stm}$ denotes the $i$-th sample in STM, and $\bm{\mathrm{s}}_{j}^{ltm}$ represents the $j$-th sample in LTM.  $n_{stm}$ and $n_{ltm}$ denote the current number of samples, respectively. When the STM pool becomes full, it is necessary to select representative samples from it and transfer them to the LTM. Hence, let $n_{stm}^{Max}$ represent the maximum number of samples that can be stored in $\mathcal{M}_{stm}$.
The sample selection process can be illustrated in \textbf{Fig. \ref{fig:fig5}} and described as follows:

\subsubsection{Relevance evaluation} 
During the inference phase of the CSC system, new samples being processed are continuously added to the STM. When the number of samples in the STM exceeds the specified maximum, an evaluation action is executed. The primary objective of this stage is to assess the relevance of samples. We evaluate the distance between two samples stored in STM and LTM using a Radial Basis Function (RBF) kernel:
\begin{equation}\label{eq:shi28}
	{\mathrm{RBF}}(\bm{\mathrm{s}}_{i}^{stm},\bm{\mathrm{s}}_{j}^{ltm})=\exp(-\frac{{\Vert \bm{\mathrm{v}}_i^{stm}-\bm{\mathrm{v}}_j^{ltm} \Vert}^2}{2 \tau ^2})
\end{equation}
where $\bm{\mathrm{v}}_i^{stm}$ and $\bm{\mathrm{v}}_j^{ltm}$ are feature vectors extracted by the semantic encoder from samples $\bm{\mathrm{s}}_{i}^{stm}$ and $\bm{\mathrm{s}}_{j}^{ltm}$, respectively. $\tau$ is the scale hyperparameter for the kernel function, and we set $\tau=10$ to ensure that the output of $\mathrm{RBF}(\cdot, \cdot)$ is within $[0, 1]$. Eq. (\ref{eq:shi28}) can be further accelerated through matrix operations, expressed as:
\begin{equation}\label{eq:shi29}
	\bm{\mathrm{S}}=\mathrm{F_{exp}}(-(\mathbf{B}^{stm}(-\mathbf{B}^{ltm})^\mathrm{T})\odot (\mathbf{B}^{stm}(-\mathbf{B}^{ltm})^\mathrm{T})/2\tau^2)
\end{equation}
where $\mathbf{B}^{stm}$ and $\mathbf{B}^{ltm}$ are feature matrices corresponding to $\mathcal{M}_{stm}$ and $\mathcal{M}_{ltm}$, respectively. $(\cdot)^\mathrm{T}$ and $\odot$ represent transpose and Hadamard product, respectively. $\mathrm{F_{exp}}(\cdot)$ is the exponential function applied element-wise to the matrix \cite{31rombach2022high}.

\subsubsection{Sample selection} The primary objective of this stage is to select samples from STM that are significantly different from those in LTM, ensuring diversity in the memory. We calculate the average similarity score between sample  $\bm{\mathrm{s}}_{i}^{stm}$ and each sample in LTM using RBF kernel:
\begin{equation}\label{eq:shi30}
	{\mathrm{R}}(\bm{\mathrm{s}}_{i}^{stm})=\frac{1}{n_{ltm}}\sum_{k=1}^{n_{ltm}}	{\mathrm{RBF}}(\bm{\mathrm{s}}_{i}^{stm},\bm{\mathrm{s}}_{k}^{ltm}).
\end{equation}

When the computed similarity score is greater than a given threshold $\lambda$, we transfer the sample from STM to LTM:
\begin{equation}\label{eq:shi31}
	{\mathrm{R}}(\bm{\mathrm{s}}_{i}^{stm})>\lambda \Rightarrow \mathcal{M}_{ltm}=\mathcal{M}_{ltm} \cup \bm{\mathrm{s}}_{i}^{stm}.
\end{equation}

After the selection is complete, $\mathcal{M}_{stm}$ 
is emptied to buffer new samples in the next round. Then, both the STM and LTM are used to train the semantic encoder and decoder through continual learning. 
The workflow of MED for the semantic encoder and decoder is illustrated in \textbf{Algorithm \ref{alg2}}.

\begin{algorithm}
	\caption{Memory-assisted Encoder and Decoder}
	\label{alg2}
	\begin{algorithmic}[1]
		\REQUIRE $\bm{\mathrm{s}},\mathcal{M}_{stm}$
		\ENSURE $\mathcal{M}_{ltm}$
		\IF{$n_{stm} \ge n_{stm}^{Max}$}
		\STATE{Calculate the kernel distance $\mathrm{RBF}(\bm{\mathrm{s}}_{i}^{stm},\bm{\mathrm{s}}_{j}^{ltm})$ between samples in STM and LTM according to Eq. (\ref{eq:shi28}).}
		\STATE{Calculate the average similarity score ${\mathrm{R}}(\bm{\mathrm{s}}_{i}^{stm})$ between sample  $\bm{\mathrm{s}}_{i}^{stm}$ and each sample in LTM according to Eq. (\ref{eq:shi30}).}	   
	   \ELSE
	   \STATE{Feed current $\bm{\mathrm{s}}$ into $\mathcal{M}_{stm}$.}
		\ENDIF
		\IF{$	{\mathrm{R}}(\bm{\mathrm{s}}_{i}^{stm})>\lambda$}
		\STATE{Transfer the $i$-th sample from $\mathcal{M}_{stm}$ to $\mathcal{M}_{ltm}$ according to Eq. (\ref{eq:shi31}).}
		\ENDIF
	\STATE{Clear $\mathcal{M}_{stm}$.}
	\end{algorithmic}
\end{algorithm}

\subsection{Noise attention module}
Inspired by the feature attention module in \cite{9xu2021wireless}, we propose a NAM based on SNR values. The NAM leverages a new noise attention network to determine the importance of each feature vector during the process of encoding and decoding, assigning weights to semantic coding and channel coding. This allows for achieving integrated encoding of both semantic and channel information according to the current SNR.

Specifically, in unfavorable channel conditions, higher weights are allocated to the channel encoder and lower weights are allocated to the semantic encoder for the same source information. This allocation strategy enhances robustness in the channel encoder to mitigate the effects of severe channel noise. Conversely, in favorable channel conditions, lower weights are assigned to the channel encoder and higher weights are assigned to the semantic encoder for the same source information. This increased allocation of weights to the semantic encoder aims to enhance semantic quality.

\begin{figure}[htpb]
	\centering
	\includegraphics[width=7cm]{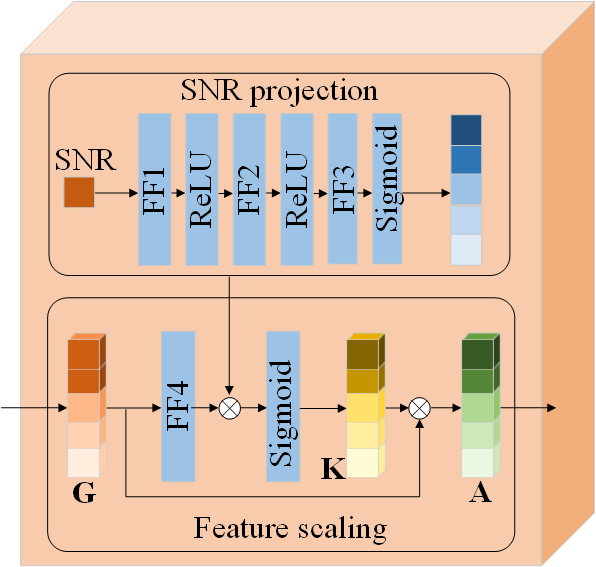}
	\caption{Noise attention module.}
	\label{fig:fig6}
\end{figure}

The structure of the NAM is illustrated in \textbf{Fig. \ref{fig:fig6}}, and a detailed description of the workflow is provided below:

\subsubsection{SNR projection} Firstly, the SNR projection module extends the SNR values to the same dimension as feature vectors in the encoder and decoder. The module is a fully connected network
comprising three FF layers. The first two FF layers employ the ReLU activation function, while the third FF layer utilizes the Sigmoid activation function.
It transforms the input SNR value $r$ to a vector $\mathbf{v}$. The mapping process from $r$ to $\mathbf{v}$ is as follows:
\begin{equation}\label{eq:shi32a}
		\bm{\mathrm{v}}'= \mathrm{ReLU}(\mathbf{W}_{n_2}\cdot \mathrm{ReLU}(\mathbf{W}_{n_1}\cdot r +b_{n_1}) +b_{n_2})
\end{equation}
\begin{equation}\label{eq:shi32c}
	\bm{\mathrm{v}}= \mathrm{Sigmoid}(\mathbf{W}_{n_3}\cdot \bm{\mathrm{v}}' +b_{n_3})
\end{equation}
where ReLU and Sigmoid denote the activation functions, and $\mathbf{W}_{n_i}$ and  $b_{n_i}$ are the weights and biases of FF layers, respectively.

\subsubsection{Feature scaling} Subsequently, we combine the input features with the projected SNR to obtain a scaling factor $\bm{\mathrm{K}}$, which records the importance of each intermediate feature vector for semantic/channel encoder and decoder as follows:
\begin{equation}\label{eq:shi33}
	\bm{\mathrm{K}}=\mathrm{Sigmoid}(\bm{\mathrm{e}}\cdot \bm{\mathrm{v}})
\end{equation}
where the Sigmoid activation function is used to constrain the output to the interval (0, 1). The $\bm{\mathrm{e}}$ is the output of the intermediate feature vectors $\bm{\mathrm{G}}$ after passing through the fourth FF layer as follows:
\begin{equation}\label{eq:shi34}
	\bm{\mathrm{e}}=\mathbf{W}_{n_4} \cdot \bm{\mathrm{G}}+b_{n_4}
\end{equation}
where $\mathbf{W}_{n_4}$ and $b_{n_4}$ are the weights and biases of the fourth FF layer.

Finally, the intermediate feature vector $\bm{\mathrm{G}}$ are multiplied by the scaling factor $\bm{\mathrm{K}}$ to obtain the calibrated vector $\bm{\mathrm{A}}$ as follows:
\begin{equation}\label{eq:shi35}
	A_i=K_i \cdot G_i
\end{equation}
where ${A}_i$ represents the $i$-th element in $\bm{\mathrm{A}}$, ${G}_i$ represents the $i$-th element in $\bm{\mathrm{G}}$, and ${K}_i$ represents the $i$-th element in $\bm{\mathrm{K}}$.
\begin{algorithm}
	\caption{Noise Attention Module}
	\label{alg3}
	\begin{algorithmic}[1]
		\REQUIRE $r, \bm{\mathrm{G}}$
		\ENSURE $\bm{\mathrm{A}}$
		\STATE{Transform the SNR value $r$ for projection and obtain $\bm{\mathrm{v}}$ according to Eqs. (\ref{eq:shi32a})-(\ref{eq:shi32c})}.
		\STATE{Transform intermediate feature vector $\bm{\mathrm{G}}$ to the vector $\bm{\mathrm{e}}$ According to Eq. (\ref{eq:shi34}).}
		\STATE{Calculate the scaling factor $\bm{\mathrm{K}}$ according to Eq. (\ref{eq:shi33}).}
		\STATE{Calculate the calibrated vector $\bm{\mathrm{A}}$ according to Eq. (\ref{eq:shi35}).}
		\STATE{Return $\bm{\mathrm{A}}$}
	\end{algorithmic}
\end{algorithm}

The NAM is embedded into the feature vectors of both the semantic/channel encoder and decoder to enhance the robustness of the CSC system. 
The workflow of NAM is illustrated in \textbf{Algorithm \ref{alg3}}.

\section{Numerical results}
In this section, we evaluate the performance of the proposed VLM-CSC system by comparing it with other SC systems.

\subsection{Simulation settings}
The datasets employed in this study include publicly available Kaggle datasets such as CIFAR, BIRDS, CATSvsDOGS, and EPPs \cite{34koehn2005europarl}. The configuration of the experiments is detailed as follows:

The pretrained BLIP has 129MB parameters, and the pretrained SD model has 1.99GB parameters. The semantic encoder comprises three transformer encoder layers alternated with NAMs. Each transformer encoder layer has 8 heads and the feature dimension is 128. The channel encoder is composed of two FF hidden layers alternating with NAMs, where the first hidden layer has 256 neurons and the second FNN layer has 128 neurons. To maintain information consistency, the semantic and channel decoder employs a structure opposite to that of the encoder. In NAM, the four FF layers have neuron quantities of 56, 128, 56, and 56, respectively. Additionally, the maximum sample size for STM is 500, and the threshold for sample selection is 0.05.

The experimental training and testing environment involves the Windows 2016 server with Python3.8, PyTorch 1.8.0 and CUDA 11.6. Computational resources are provided by an Intel(R) Xeon(R) Silver 4210R CPU @ 2.40GHz and NVIDIA Tesla T4.

\subsection{Evaluation metrics}
The proposed VLM-CSC system transforms image data to textual semantic data through the BLIP-based knowledge base, encodes it using a semantic encoder, decodes it at the receiver, and finally reconstructs the image through the SD-based knowledge base. To assess the performance of the VLM-CSC system, two corresponding metrics are designed: (1) Image-level, examining the accuracy of semantic reconstruction for image data; (2) Text-level, examining the accuracy of semantic recovery for text data.

\subsubsection{Image-level: Semantic Service Quality (SSQ)} In performance assessment of the SC system, the emphasis on semantic layer transmission should be directed towards whether information, after undergoing semantic recovery, can meet the expectations of subsequent tasks. The general quality metric for semantic services is denoted by \cite{45dong2022semantic}:
\begin{equation}\label{eq:shi24}
	SSQ=\frac{ST(\hat{S})}{ST(S)}
\end{equation}
where \(S\) represents the unprocessed source information at the transmitter, \(\hat{S}\) represents the recovered information at the semantic level by the receiver, and \(ST(\cdot)\) signifies the performance of the source information or recovered information when executing subsequent tasks, which is the classification accuracy in our study.

\subsubsection{Text-level: Bilingual Evaluation Understudy (BLEU)} The BLEU score outputs a number between 0 and 1, indicating how similar the decoded text is to the transmitted text, with 1 representing the highest similarity. For a transmission sentence $\bm{\mathrm{s}}$ with length $l_s$ and a decoded sentence $\bm{\mathrm{\hat{s}}}$ with length $l_{\hat{s}}$, BLEU can be expressed as \cite{37xie2021deep}:
\begin{equation}\label{eq:shi6}
	\log {\mathrm{BLEU}}=\min (1-\frac{l_{\hat{s}}}{l_s},0)+\sum_{n=1}^N u_n\log p_n
\end{equation}
where the "n-gram" refers to a contiguous sequence of $n$ words from a given sample of text or speech, $u_n$ is the weight of the $n$-grams, and $p_n$ is the $n$-grams score, defined as:
\begin{equation}\label{eq:shi7}
	p_n=\frac{\sum_k \min ({C_k}(\mathbf{\hat{s}}),C_k(\bm{\mathrm{s}}))}{\sum_k\min({C_k}(\bm{\mathrm{\hat{s}}}))}
\end{equation}
where $C_k(\cdot)$ is the frequency count function for the $k$-th element in the $n$-th grams.

\subsection{Performance comparison of VLM-base KBs}
\begin{figure}[htpb]
	\centering
	\includegraphics[width=8cm]{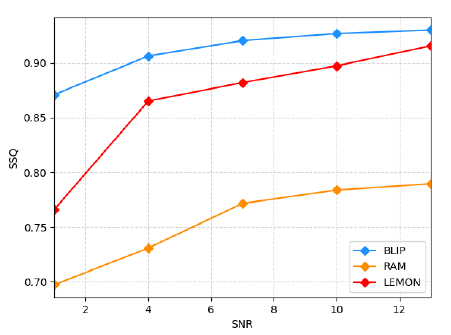}
	\caption{SSQ of CSC systems based on different VLMs.}
	\label{fig:fig8}
\end{figure}

To evaluate the performance of extracting semantic information from images using KBs, we employ three VLMs (BLIP, LEMON\cite{40hu2022scaling}, and RAM\cite{41zhang2023recognize}) to construct the sender-side KBs in the CSC system. The receiver-side KB is uniformly implemented using the SD model. Subsequently, we assess the CSC system's performance on the AWGN channel. SSQ is utilized as the evaluation metric on the CATSvsDOGS dataset \cite{34koehn2005europarl}.
The experimental outcomes are illustrated in \textbf{Fig. \ref{fig:fig8}}.

From \textbf{Fig. \ref{fig:fig8}}, it is evident that the CSC system based on BLIP exhibits the highest SSQ, followed by the one based on LEMON, while the CSC system based on RAM performs the poorest, significantly lower than the CSC systems based on BLIP and LEMON. Furthermore, the CSC system based on BLIP maintains robust performance even at low SNR values. The experimental results indicate that the CSC system constructed based on BLIP accurately extracts image semantics and sustains commendable performance across different SNR levels.

\subsection{Performance evaluation for MED}
\begin{figure*}[htpb]
	\centering
	\includegraphics[width=19cm]{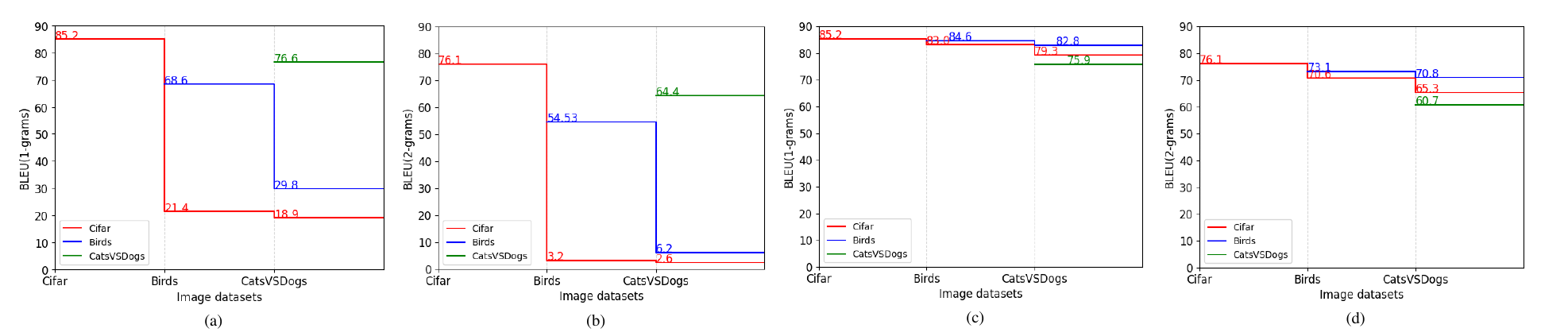}
	\caption{The continual learning map for BLEU scores across diverse image datasets are evaluated in the following scenarios:(a) The BLEU (1-grams) of VLM-CSC without MED across different image datasets. (b) The BLEU (2-grams) of VLM-CSC without MED across different image datasets. (c) The BLEU (1-grams) of VLM-CSC across different image datasets. (d) The BLEU (2-grams) of VLM-CSC across different image datasets.}
	\label{fig:fig9}
\end{figure*}

To demonstrate the performance of the proposed MED, we conduct experiments comparing VLM-CSC with the MED module against VLM-CSC without the MED module. The evaluation is performed across different image datasets. The image datasets include Cifar, Birds, and CatsVSDogs \cite{34koehn2005europarl}.
BLEU scores for semantic similarity serve as the evaluation metric. Additionally, when assessing the performance of VLM-CSC on image datasets with different distributions, the channel is fixed to Rayleigh. 
The continual learning map, originally proposed by Google, is employed to visualize the performance changes of existing tasks when a new task is introduced.
The experimental results are illustrated by the continual learning map in \textbf{Fig. \ref{fig:fig9}}.

Figure \textbf{Fig. \ref{fig:fig9}} (a) and  (b) illustrate a significant performance drop in the VLM-CSC system without the MED module on the previous Cifar dataset after learning subsequent datasets such as Birds and CatsVSDogs. In contrast, \textbf{Fig. \ref{fig:fig9}} (c) and (d) reveal that the VLM-CSC system with the MED module only exhibits a marginal decline in performance on the previous Cifar dataset after learning subsequent datasets like Birds and CatsVSDogs. 

The experimental results from \textbf{Fig. \ref{fig:fig9}} underscore that the proposed MED module enables the CSC system to overcome catastrophic forgetting during the continual learning process. This facilitates knowledge learning from multiple image datasets, enhancing the generalization of the CSC system in dynamic environments.

\subsection{Performance evaluation for NAM}
\begin{figure}[htpb]
	\centering
	\includegraphics[width=8cm]{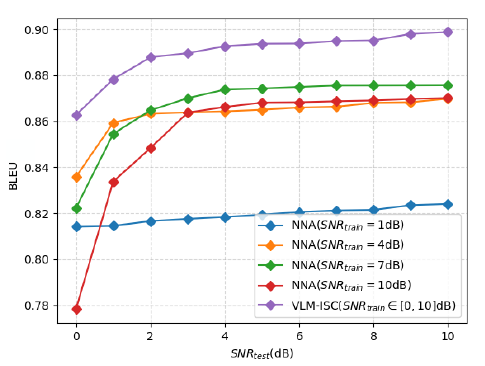}
	\caption{The performance of NAM in the VLM-CSC system. NNA represents the VLM-CSC system without NAM.}
	\label{fig:fig10}
\end{figure}

To demonstrate the performance of the proposed  NAM, we conduct an experimental comparison between VLM-CSC with and without NAM. Semantic similarity, measured by BLEU score, serves as the evaluation metric. Specifically, the proposed VLM-CSC system is trained under a uniform distribution of $SNR_{train}$ ranging from 0 dB to 10 dB, while the VLM-CSC system without NAM is trained at specific $SNR_{train}$ values of 1 dB, 4 dB, 7 dB, and 10 dB. Subsequently, the performance of the VLM-CSC system is evaluated at specific $SNR_{test}$ values ranging from 0 dB to 10 dB. The experimental results are depicted in \textbf{Fig. \ref{fig:fig10}}.

The findings depicted in Figure \ref{fig:fig10} demonstrate that the performance of the proposed VLM-CSC system outperforms any VLM-CSC system without NAM, specifically trained at distinct $SNR_{train}$ values. This observation highlights the capability of the VLM-CSC system, equipped with NAM, to address the performance degradation challenges caused by the mismatch between the SNR during training and deployment stages in conventional ISC systems. This improvement contributes to the robustness of the VLM-CSC system across different SNR values.

\subsection{Semantic communication performance evaluation}
\begin{figure*}[htpb]
	\centering
	\includegraphics[width=19cm]{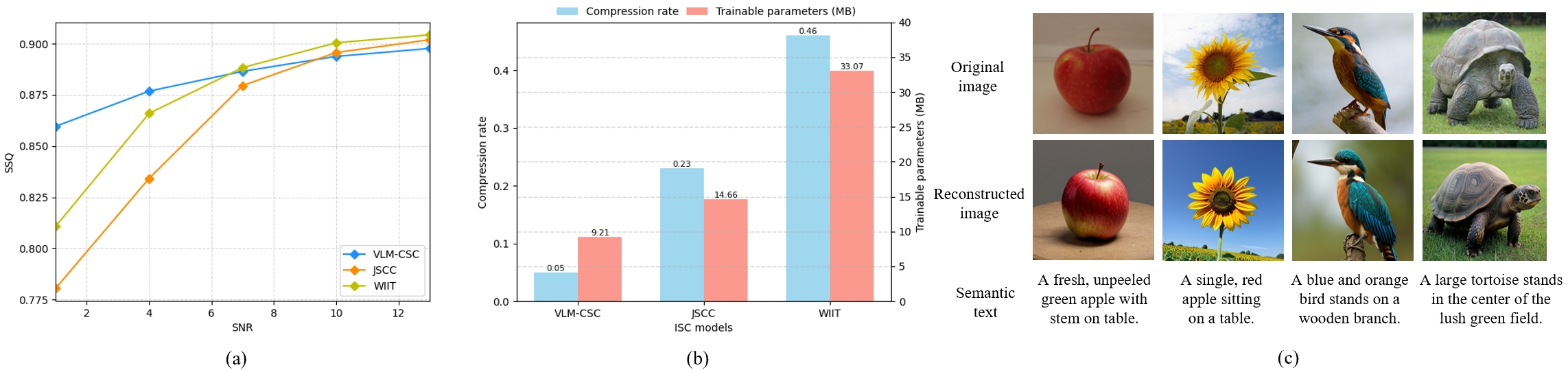}
	\caption{Performance comparison of VLM-CSC with other ISC systems. (a) SSQ. (b) Compression ratio and trainable parameters. (c) Semantic alignment.}
	\label{fig:fig7}
\end{figure*}

To evaluate the performance of the VLM-CSC system in image classification tasks, we compare it with JSCC based on CNN \cite{38kurka2020deepjscc} and WITT based on ViT \cite{39yang2023witt}. The metric used for performance evaluation is classification accuracy. Additionally, we assess the bandwidth-saving capabilities of VLM-CSC by considering the compression ratio between transmitted data and original images as the evaluation metric. The experimental results are presented in \textbf{Fig. \ref{fig:fig7}}.

\textbf{Fig. \ref{fig:fig7}} (a) clearly demonstrates that, at low SNR levels, the superior performance of VLM-CSC in the classification task with the CATSvsDOGS dataset, and WITT shows slightly lower results, particularly with decreased performance compared to VLM-CSC. At high SNR levels, WIIT and JSCC exhibit superior SSQ compared to VLM-CSC due to their direct transmission of images. \textbf{Fig. \ref{fig:fig7}} (b) depicts the compression ratio and trainable parameters, with VLM-CSC achieving the lowest of all, followed by JSCC, while WITT attains the highest compression ratio and trainable parameters. \textbf{Fig. \ref{fig:fig7}} (c) illustrates that the reconstructed image highly aligns with the original image and the image description, validating the VLM-CSC system's ability to ensure semantic consistency across modalities.

The experimental results depicted in \textbf{Fig. \ref{fig:fig7}} demonstrate that the proposed VLM-CSC exhibits overall superior performance in image classification tasks compared to other ISC systems at low SNR levels. Then, the compression ratio of transmitted data is significantly lower for VLM-CSC compared to other ISC systems, indicating that VLM-CSC can effectively conserve transmission bandwidth while preserving high-quality semantic transmission. Moreover, due to the absence of training VLMs, the VLM-CSC system exhibits the minimum number of trainable parameters, resulting in the lowest training complexity.

\section{Conclusion}
This paper introduces a novel VLM-CSC system capable of converting images into text descriptions for transmission over wireless channels, and reconstructing the image at the receiver. The system includes three main contributions: CKB for image-to-text and text-to-image conversion, MED for continual learning in dynamic environments, and NAM for joint semantic and channel encoding based on SNR. Corresponding performance metrics are designed to evaluate the VLM-CSC system from both image and text perspectives. Experimental validations are conducted under various image datasets. Results demonstrate the effectiveness and robustness of the VLM-CSC system in preserving semantic similarity between the image and text, as well as its adaptability to dynamic environments.


\bibliographystyle{ieeetran}
\bibliography{bare_jrnl_bobo}
\newpage
\end{document}